\documentclass{article}
\usepackage{spconf,amsmath,graphicx,hyperref}

\usepackage{latexsym}
\usepackage{amssymb}
\usepackage{amsmath}
\usepackage{amsthm}
\usepackage{booktabs}
\usepackage{enumitem}
\usepackage{graphicx}
\usepackage{color}

\usepackage{cite}
\usepackage{amsmath,amssymb,amsfonts}
\usepackage{algorithmic}
\usepackage{graphicx}
\usepackage{hyperref}
\usepackage{cleveref}
\usepackage{subcaption}
\usepackage{makecell}
\usepackage{tabularx}
\usepackage{multirow}
\usepackage{textcomp}
\usepackage{xcolor}
\usepackage{makecell}

\title{OUT-OF-DISTRIBUTION DETECTION BASED ON TOTAL VARIATION ESTIMATION}
\name{Dabiao Ma$^{\ast}$, Zhiba Su$^{\ast}$, Jian Yang, Haojun Fei$^{\dagger}$\thanks{$^{\ast}$Equal contribution. $^{\dagger}$Corresponding author.}}
\address{Qifu Technology \\
\small \texttt{\{madabiao-jk, wangye3-jk, zhangchulan-jk\}@qifu.com, zhiba.su.1995@gmail.com}}
\begin{document}
%
\maketitle
\begin{abstract}
This paper introduces a novel approach to securing machine learning model deployments against potential distribution shifts in practical applications, the Total Variation Out-of-Distribution (TV-OOD) detection method. Existing methods have produced satisfactory results, but TV-OOD improves upon these by leveraging the Total Variation Network Estimator to calculate each input's contribution to the overall total variation. By defining this as the total variation score, TV-OOD discriminates between in- and out-of-distribution data. The method's efficacy was tested across a range of models and datasets, consistently yielding results in image classification tasks that were either comparable or superior to those achieved by leading-edge out-of-distribution detection techniques across all evaluation metrics.
\end{abstract}
\section{Introduction}
\label{sec:intro}

Deep Neural Networks (DNNs) are crucial to many modern applications, but they often face challenges when dealing with inputs that differ from their training data, such as noisy, corrupted, or entirely new concepts. This can lead to incorrect predictions due to their overconfidence bias. Thus, distinguishing between out-of-distribution (OOD) and in-distribution (ID) data is vital.

\cite{ontheusefullness} critiques mutual information for OOD detection, highlighting its underperformance and methodological flaws. Their approach calculates mutual information for ID images and extends it to OOD images, but this is invalid as the function is not defined for OOD data. In contrast, our Total Variation Out-of-Distribution(TV-OOD) method introduces a robust score based on total variation, incorporating both ID and OOD data with an additional fake label. This approach, outperforming KL divergence in ablation studies, sets new benchmarks for OOD detection in image classification.

\section{Related Work}
\label{sec:related}

Effective OOD detection relies on a suitable score function, with methods categorized by their approach. Output-based methods utilize classifier outputs, such as MSP\cite{baseline}, ODIN\cite{Odin} with temperature scaling, OE\cite{anomaly} with auxiliary data, Energy\cite{energy}, and Balanced EnergyOE\cite{balancedEnergy} addressing label imbalance. Feature-based methods analyze features, including Mahalanobis\cite{unified} distance, RankFeat\cite{RankFeat} for reducing overconfidence, KNN\cite{knn}, and VIM\cite{vim} combining feature and logit scores. Gradient-based methods like GradNorm\cite{gradnorm} use gradient norms derived from backpropagation to distinguish OOD data. Activation-based methods improve detection by correcting activations, such as clipping (ReAct\cite{react}), filtering (BFAct\cite{BFAct}), or percentile-based nullification (ASH-S\cite{ASH}).

\section{Preliminaries}
\subsection{Problem Statement}

OOD detection involves categorizing an input as ID or OOD, essentially making it a binary classification problem. Consider a well-trained classifier, $\operatorname{f}$, trained on an in-distribution $P_{in}$. Any data distribution beyond this training set is the out-of-distribution, $P_{out}$. The support of $P_{in}$ and $P_{out}$ are disjoint. For an input $x$, we aim to create a binary estimator:
\begin{equation}
\begin{aligned}
    \operatorname{g}(x) =
    \begin{cases}
        in & \text{if } \operatorname{S}(x) \geq \tau \\
        out & \text{if } \operatorname{S}(x) < \tau
    \end{cases}
\end{aligned}    
\end{equation}
Here, $\operatorname{S}(x)$ is the score function, and threshold $\tau$ ensures a significant portion (e.g., 95\%) of ID data is correctly identified. While various score function designs exist, our method uniquely employs total variation estimation.

\subsection{MINE}

Mutual information network estimator(MINE)\cite{mine} is a trainable neural network that can estimate the mutual information between two random variables $X$ and $Y$:
\begin{equation}
\label{equ:1}
    \begin{aligned}
        \operatorname{I}(X,Y)=\operatorname{D_{KL}} \left(P_{XY}||P_X \cdot P_Y \right) 
    \end{aligned}
\end{equation}
Where the KL divergence $\operatorname{D_{KL}}$ is defined as:
\begin{equation}
    \begin{aligned}
         \operatorname{D_{KL}} \left( p||q \right) = \mathbb{E}_p \left[ \log\frac{dp}{dq} \right]
    \end{aligned}
\end{equation}
MINE approximates mutual information using neural networks and the Donsker-Varadhan representation of KL divergence:
\begin{equation}
\label{equ:2}
    \begin{aligned}
         \operatorname{D_{KL}} \left( p||q \right) =\sup_{\operatorname{T}:\omega \mapsto \mathbb{R}} \mathbb{E}_p \left[ \operatorname{T} \right] - \log \left(\mathbb{E}_q \left[ e^{\operatorname{T}} \right]  \right)
    \end{aligned}
\end{equation}
Where the supremum is taken over all functions $\operatorname{T}$ such that the two expectations are finite. Combined with \cref{equ:1}, MINE is defined as:
\begin{equation}
\label{equ:3}
    \begin{split}
         \sup_\theta \mathbb{E}_{x,y \sim P_{XY}}\left[ \operatorname{T_{\theta}(x,y)} \right] - \\ 
         \log \left(
           \mathbb{E}_{x' \sim P_{X}, y' \sim P_{Y}} \left[ e^{\operatorname{T_{\theta}(x',y')}}\right] 
               \right)
    \end{split}
\end{equation}
Where $\operatorname{T_{\theta}}$ is a neural network with trainable parameters $\theta$.

\section{Method: TV-OOD}

Within this section, we focus on a $K$-class model. Our approach introduces an original Total Variation Network Estimator (TVNE) to analyze the dependence between two random variables: X, sourced from the distribution of the comprehensive dataset, $X \sim P_X$, where the support of $P_X$ is the amalgamation of the supports of $P_{in}$ and $P_{out}$, and Y, which assigns $j \in [1, K]$ when drawn from in-distribution $P_{in}$ or $K+1$ when sourced from out-of-distribution $P_{out}$.


The KL divergence in MINE is a specific case within the f-divergence family. Here, we modify MINE by introducing total variation distance, another f-divergence form, as an alternative measure of disparity. Its variational representation is given as follows.

\begin{equation}
\label{equ:4}
    \begin{aligned}
         \operatorname{D_{TV}} \left( p||q \right) =\sup_{\left| \operatorname{T} \right| \leq \frac{1}{2}} \mathbb{E}_p \left[ \operatorname{T} \right] - \mathbb{E}_q \left[ \operatorname{T} \right]
    \end{aligned}
\end{equation}

The relationship between total variation distance and KL divergence is clarified by the Bretagnolle–Huber inequality, which suggests that $D_{TV} \leq \sqrt{1-exp\left( -D_{KL}\right)}$. This suggests that as $D_{TV} \to 1$, $D_{KL} \to \infty$, indicating that maximizing the estimation of total variation distance concurrently maximizes the estimation of mutual information to a certain extent.

Total variation distance has two advantages over KL divergence. First, the $ \log \left(\mathbb{E}_q \left[ e^{\operatorname{T}} \right] \right) $ term in KL divergence’s Donsker-Varadhan representation requires expectations over the entire dataset, but its mini-batch implementation introduces bias, only partially reduced by an exponential moving average. The variational representation of total variation avoids this by enabling linear separation of terms into mini-batches. Second, Pinsker's Inequality, which suggests that $D_{TV}^2 \leq D_{KL}/2$, indicates that $D_{KL}$ is weaker than $D_{TV}$, and that $D_{TV}$ is a more sensitive metric\cite{GhaziI24}. This is confirmed by our ablation study in \cref{sec:f-d}, which shows that total variation outperforms other f-divergences, including KL divergence, further establishing its superiority.


By inserting $P_X$, $P_Y$ and $P_{XY}$ into \cref{equ:4}, we derive TVNE:
\begin{equation}
\label{equ:5}
    \begin{aligned}
         \sup_{\theta, \left| \operatorname{T_{\theta}} \right| \leq \frac{1}{2}} \mathbb{E}_{x,y \sim P_{XY}}\left[ \operatorname{T_{\theta}(x,y)} \right] \\ - 
           \mathbb{E}_{x' \sim P_{X}, y' \sim P_{Y}} \left[ \operatorname{T_{\theta}(x',y')}\right]      
    \end{aligned}
\end{equation}
Using the definition of $P_Y$, \cref{equ:5} is further transformed into:
\begin{equation}
\label{equ:6}
    \begin{aligned}
         & \sup_{\theta, \left| \operatorname{T_{\theta}} \right| \leq \frac{1}{2}} \mathbb{E}_{x,y \sim P_{XY}}\left[ \operatorname{T_{\theta}(x,y)} \right] \\
         & - m\sum_{j=1}^{K}m_j\mathbb{E}_{x \sim P_{X}} \left[ \operatorname{T_{\theta}}(x,j)\right]  \\
         & -(1-m)\mathbb{E}_{x \sim P_{X}} \left[ \operatorname{T_{\theta}}(x,K+1)\right]        
    \end{aligned}
\end{equation}
\begin{figure*}[htbp]
  \centering
  \includegraphics[width=\textwidth]{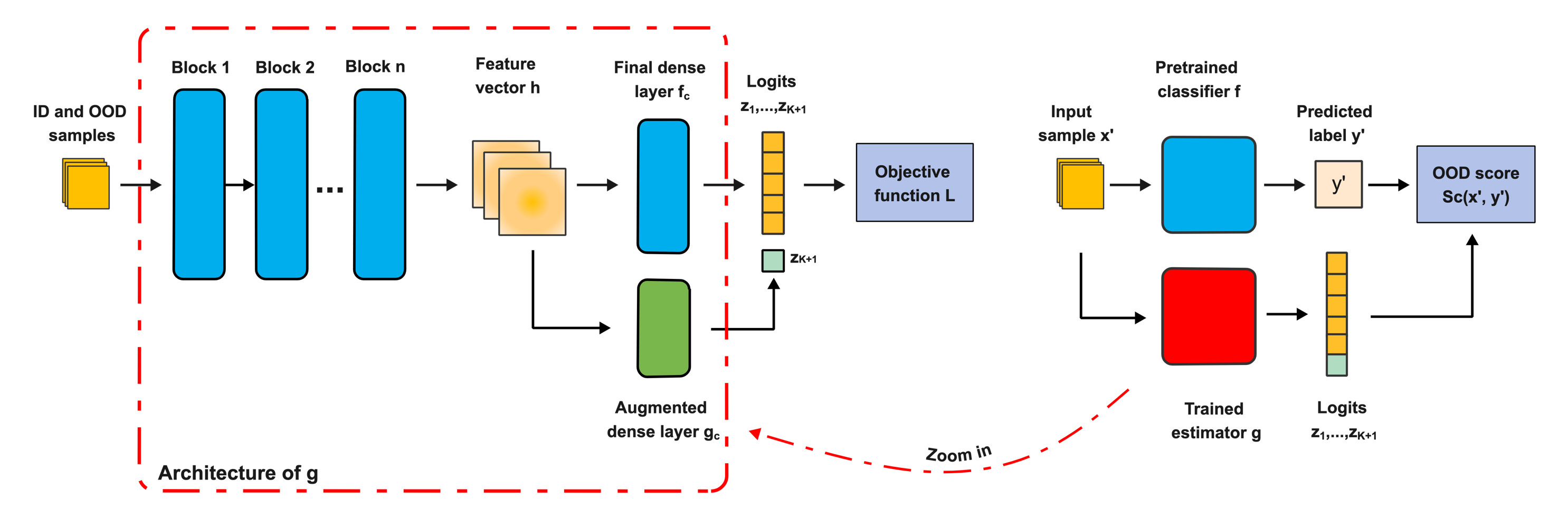}
  \caption{Schematic of TV-OOD implementation: Blue blocks represent the pre-trained classifier $\operatorname{f}$, while red dashed lines/blocks denote the total variation neural estimator $\operatorname{g}(x)$, integrated via an auxiliary dense layer $\operatorname{g_c}(h)$. The left shows training, and the right depicts inference. Both the objective $L$ and OOD score $\operatorname{Sc}$ derive from \cref{equ:7}.}
  
  \label{fig:both_images}
\end{figure*}
where $m$ signifies the ratio of ID samples to the aggregate number of ID and OOD samples, and $m_j$ signifies the ratio of ID samples belonging to a specific class $j$ to the aggregate number of ID samples. $m$ determines the relative influence of the terms $\operatorname{T_{\theta}}(x,j), j \in [1, K]$ and $\operatorname{T_{\theta}}(x,K+1)$. We determine $m_j$ from the training dataset, leaving $m$ as a tunable hyperparameter.

Given a set of $N$ training samples $(x_1,y_1), \ldots, (x_N,y_N)$, we infer from \cref{equ:6} that TVNE seeks to maximize the following objective function $L$:
\begin{equation}
\label{equ:7}
    \begin{aligned}
         L= & \frac{1}{N} \sum_{i=1}^{N} \left[ 
         \begin{aligned}
         & \operatorname{T_{\theta}}(x_i,y_i) - m\sum_{j=1}^{K}m_j\operatorname{T_{\theta}}(x_i,j) \\ & -(1-m)\operatorname{T_{\theta}}(x_i,K+1)
         \end{aligned}
         \right] \\ =& \frac{1}{N} \sum_{i=1}^{N} \operatorname{Sc}(x_i,y_i)    
    \end{aligned}
\end{equation}
We define $\operatorname{Sc}(x,y)$ as the score function of input $x$ in the TV-OOD method, with $y$ denoting the classifier's predicted label. Next, we assess the effectiveness of $\operatorname{Sc}$ as an OOD score for OOD detection. During inference, when the input $x$ is sourced from the in-distribution, the $\operatorname{Sc}(x,y)$ score naturally achieves a high value due to the maximization of the objective function $L$. Conversely, when the input $x$ is sourced from OOD but mistakenly classified as one of the in-distribution classes by the classifier, a simple calculation can derive the expected $\operatorname{Sc}$ score of $x$:
\begin{equation}
\label{equ:8}
    \begin{aligned}
         \overline{\operatorname{Sc}(x)} & = \sum_{j'=1}^{K}m_{j'}\operatorname{Sc}(x,j') \\
         & = -\frac{1-m}{m} \operatorname{Sc}(x,K+1)
    \end{aligned}
\end{equation}

As a result, the expected $\operatorname{Sc}$ score of $x$, when $x$ is sampled from OOD, equals the inverse of $\operatorname{Sc}(x,K+1)$ multiplied by a constant. Maximizing the objective function $L$ leads to a high value of $\operatorname{Sc}(x,K+1)$, which consequently reduces the expected $\operatorname{Sc}$ score of $x$. This analysis highlights the link between TVNE and OOD detection, leading to the deduction of the score function $\operatorname{Sc}(x,y)$. Accordingly, we propose TV-OOD, utilizing $\operatorname{Sc}(x,y)$ as its score. The comprehensive process of our TV-OOD method is visually portrayed in \cref{fig:both_images}. 

\begin{figure}[htbp]
\centering
\begin{minipage}[b]{\linewidth}
\centering
\centerline{\includegraphics[width=7.0cm]{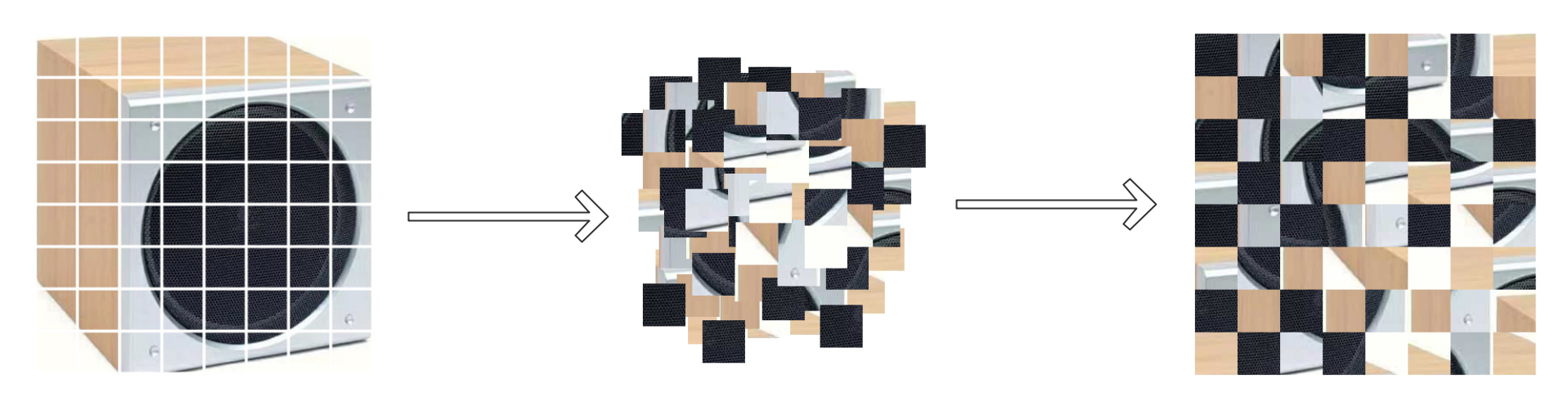}}
\caption{Uniform-sized sub-blocks are extracted from the image. The sub-blocks are sized to be $\frac{1}{64}$ of the original image.}
\label{fig:aug}
\end{minipage}
\end{figure}

Our training approach utilizes an auxiliary OOD dataset, As shown in \cref{fig:aug}, our data augmentation divides an image into uniform sub-blocks, rearranging them into a new image unrecognizable as any ID category. Using $D_{aug}$ with auxiliary OOD datasets can optimize our method. Preliminary evaluations of techniques like AugMix\cite{augmix}, CutMix\cite{cutmix}, and synthetic negatives\cite{gligen} are included in ablation studies.

\section{Experiments}

\begin{table*}[ht]
    \centering
    \caption{Subtables (a)–(d) report performance on CIFAR-100 and ImageNet-1k as $D_{in}$ across architectures. Each entry shows the average over $D_{out}^{test}$ datasets. $\downarrow$ indicates lower values are better, and $\uparrow$ higher. TV-OOD uses $D^{train}_{out}$ and $D_{aug}$ when compared to methods requiring auxiliary OOD datasets.}
    \label{table:1}
    \scriptsize 
\vspace{0.5cm}
\begin{minipage}{.46\linewidth}
        \centering

        \subcaption{Results on DenseNet121 using CIFAR-100 as $D_{in}$.}
        \fontsize{6}{7.2}\selectfont{
\begin{tabularx}{\linewidth}{XX|XXX}
\hline
\multicolumn{2}{c|}{\centering  $D_{in}$~+~model} &\multicolumn{3}{c}{\centering  cifar100~+~DenseNet121}    \\
\hline 
\multicolumn{2}{c|}{} & \multicolumn{1}{c}{ FPR95$\downarrow$} & \multicolumn{1}{c}{AUROC$\uparrow$} & \multicolumn{1}{c}{ AUPR$\uparrow$} \\
\hline
\multicolumn{1}{c}{\multirow{9}{*}{\makecell{w/o \\ auxiliary \\ OOD \\ dataset}}} & \multicolumn{1}{c|}{MSP} & \multicolumn{1}{c}{78.55} & \multicolumn{1}{c}{78.34} & \multicolumn{1}{c}{94.45}  \\
& \multicolumn{1}{c|}{Energy} & \multicolumn{1}{c}{74.37} & \multicolumn{1}{c}{80.18} & \multicolumn{1}{c}{94.91}  \\
& \multicolumn{1}{c|}{ODIN} & \multicolumn{1}{c}{74.06} & \multicolumn{1}{c}{80.14} & \multicolumn{1}{c}{94.93}  \\
& \multicolumn{1}{c|}{Mahalanobis} & \multicolumn{1}{c}{34.16} & \multicolumn{1}{c}{88.24} & \multicolumn{1}{c}{96.76} \\
& \multicolumn{1}{c|}{LogitNorm} & \multicolumn{1}{c}{72.42} & \multicolumn{1}{c}{81.71} & \multicolumn{1}{c}{94.20} \\
& \multicolumn{1}{c|}{KNN} & \multicolumn{1}{c}{36.57} & \multicolumn{1}{c}{91.56} & \multicolumn{1}{c}{97.88} \\
& \multicolumn{1}{c|}{ASH-S} & \multicolumn{1}{c}{62.19} & \multicolumn{1}{c}{83.79} & \multicolumn{1}{c}{96.13} \\
& \multicolumn{1}{c|}{VIM} & \multicolumn{1}{c}{\textbf{28.70}} & \multicolumn{1}{c}{\textbf{92.45}} & \multicolumn{1}{c}{\textbf{98.08}} \\
& \multicolumn{1}{c|}{TV-OOD} & \multicolumn{1}{c}{46.01} & \multicolumn{1}{c}{88.34} & \multicolumn{1}{c}{97.05} \\
\hline
\multicolumn{1}{c}{\multirow{5}{*}{\makecell{with \\ auxiliary \\ OOD \\ dataset}}} & \multicolumn{1}{c|}{OE} & \multicolumn{1}{c}{64.93} & \multicolumn{1}{c}{85.71} & \multicolumn{1}{c}{96.68}  \\
& \multicolumn{1}{c|}{EF} & \multicolumn{1}{c}{59.94} & \multicolumn{1}{c}{86.14} & \multicolumn{1}{c}{96.68} \\
& \multicolumn{1}{c|}{WOODS} & \multicolumn{1}{c}{54.99} & \multicolumn{1}{c}{86.02} & \multicolumn{1}{c}{94.35}  \\
& \multicolumn{1}{c|}{BEOE} & \multicolumn{1}{c}{47.58} & \multicolumn{1}{c}{91.35} & \multicolumn{1}{c}{98.01}  \\
& \multicolumn{1}{c|}{TV-OOD} & \multicolumn{1}{c}{\textbf{24.93}} & \multicolumn{1}{c}{\textbf{93.87}} & \multicolumn{1}{c}{\textbf{98.42}}  \\
\hline
\end{tabularx}
\par }
\end{minipage}\hfill
\begin{minipage}{.46\linewidth}
    \subcaption{Results on WideResNet using CIFAR-100 as $D_{in}$.}
     \fontsize{6}{7.2}\selectfont{\begin{tabularx}{\linewidth}{XX|XXX}

\hline
\multicolumn{2}{c|}{\centering  $D_{in}$~+~model} &\multicolumn{3}{c}{\centering  cifar100~+~WideResNet}    \\
\hline 
\multicolumn{2}{c|}{} & \multicolumn{1}{c}{ FPR95$\downarrow$} & \multicolumn{1}{c}{AUROC$\uparrow$} & \multicolumn{1}{c}{ AUPR$\uparrow$} \\
\hline
\multicolumn{1}{c}{\multirow{9}{*}{\makecell{w/o \\ auxiliary \\ OOD \\ dataset}}} & \multicolumn{1}{c|}{MSP} & \multicolumn{1}{c}{80.60} & \multicolumn{1}{c}{75.06} & \multicolumn{1}{c}{93.74}  \\
& \multicolumn{1}{c|}{Energy} & \multicolumn{1}{c}{74.12} & \multicolumn{1}{c}{79.02} & \multicolumn{1}{c}{94.69}  \\
& \multicolumn{1}{c|}{ODIN} & \multicolumn{1}{c}{74.09} & \multicolumn{1}{c}{78.97} & \multicolumn{1}{c}{94.66}   \\
& \multicolumn{1}{c|}{Mahalanobis} & \multicolumn{1}{c}{57.01} & \multicolumn{1}{c}{82.22} & \multicolumn{1}{c}{95.38}\\
& \multicolumn{1}{c|}{LogitNorm} & \multicolumn{1}{c}{56.12} & \multicolumn{1}{c}{86.6} & \multicolumn{1}{c}{96.61} \\
& \multicolumn{1}{c|}{KNN} & \multicolumn{1}{c}{55.02} & \multicolumn{1}{c}{85.25} & \multicolumn{1}{c}{95.16} \\
& \multicolumn{1}{c|}{ASH-S} & \multicolumn{1}{c}{53.32} & \multicolumn{1}{c}{84.56} & \multicolumn{1}{c}{95.96} \\
& \multicolumn{1}{c|}{VIM} & \multicolumn{1}{c}{51.53} & \multicolumn{1}{c}{86.55} & \multicolumn{1}{c}{96.59} \\
& \multicolumn{1}{c|}{TV-OOD} & \multicolumn{1}{c}{\textbf{41.58}} & \multicolumn{1}{c}{\textbf{88.12}} & \multicolumn{1}{c}{\textbf{96.91}} \\
\hline
\multicolumn{1}{c}{\multirow{5}{*}{\makecell{with \\ auxiliary \\ OOD \\ dataset}}} & \multicolumn{1}{c|}{OE} & \multicolumn{1}{c}{57.57} & \multicolumn{1}{c}{86.65} & \multicolumn{1}{c}{96.82}  \\
& \multicolumn{1}{c|}{EF} &  \multicolumn{1}{c}{48.99} & \multicolumn{1}{c}{87.63} & \multicolumn{1}{c}{96.93}  \\
& \multicolumn{1}{c|}{WOODS} & \multicolumn{1}{c}{43.49} & \multicolumn{1}{c}{87.16} & \multicolumn{1}{c}{96.05}  \\
& \multicolumn{1}{c|}{BEOE} & \multicolumn{1}{c}{39.08} & \multicolumn{1}{c}{91.34} & \multicolumn{1}{c}{\textbf{97.98}}  \\
& \multicolumn{1}{c|}{TV-OOD} & \multicolumn{1}{c}{\textbf{27.39}} & \multicolumn{1}{c}{\textbf{92.49}} & \multicolumn{1}{c}{97.91}  \\
\hline
\end{tabularx}
\par }
\end{minipage}\hfill
\begin{minipage}{.46\linewidth}
\centering
\vspace{1cm}
\subcaption{Results on ViT-B\_16 using CIFAR-100 as $D_{in}$.}
\fontsize{6}{7.2}\selectfont{
\begin{tabularx}{\columnwidth}{XX|XXX}
\hline
\multicolumn{2}{c|}{\centering  $D_{in}$~+~model} &\multicolumn{3}{c}{\centering  cifar100~+~ViT-B\_16}    \\
\hline 
\multicolumn{2}{c|}{} & \multicolumn{1}{c}{ FPR95$\downarrow$} & \multicolumn{1}{c}{AUROC$\uparrow$} & \multicolumn{1}{c}{ AUPR$\uparrow$} \\
\hline
\multicolumn{1}{c}{\multirow{9}{*}{\makecell{w/o \\ auxiliary \\ OOD \\ dataset}}} & \multicolumn{1}{c|}{MSP} & \multicolumn{1}{c}{44.91} & \multicolumn{1}{c}{91.93} & \multicolumn{1}{c}{98.22}  \\
& \multicolumn{1}{c|}{Energy} & \multicolumn{1}{c}{24.24} & \multicolumn{1}{c}{94.95} & \multicolumn{1}{c}{98.83} \\
& \multicolumn{1}{c|}{ODIN} & \multicolumn{1}{c}{25.66} & \multicolumn{1}{c}{94.79} & \multicolumn{1}{c}{98.79}  \\
& \multicolumn{1}{c|}{Mahalanobis} & \multicolumn{1}{c}{20.49} & \multicolumn{1}{c}{\textbf{96.18}} & \multicolumn{1}{c}{\textbf{99.19}} \\
& \multicolumn{1}{c|}{LogitNorm} & \multicolumn{1}{c}{24.56} & \multicolumn{1}{c}{94.53} & \multicolumn{1}{c}{98.59} \\
& \multicolumn{1}{c|}{KNN} & \multicolumn{1}{c}{31.66} & \multicolumn{1}{c}{93.03} & \multicolumn{1}{c}{98.41} \\
& \multicolumn{1}{c|}{ASH-S} & \multicolumn{1}{c}{25.90} & \multicolumn{1}{c}{94.48} & \multicolumn{1}{c}{98.69} \\
& \multicolumn{1}{c|}{VIM} & \multicolumn{1}{c}{22.66} & \multicolumn{1}{c}{93.99} & \multicolumn{1}{c}{98.78} \\
& \multicolumn{1}{c|}{TV-OOD} & \multicolumn{1}{c}{\textbf{20.48}} & \multicolumn{1}{c}{94.88} & \multicolumn{1}{c}{98.63} \\
\hline
\multicolumn{1}{c}{\multirow{5}{*}{\makecell{with \\ auxiliary \\ OOD \\ dataset}}} & \multicolumn{1}{c|}{OE} & \multicolumn{1}{c}{14.62} & \multicolumn{1}{c}{95.71} & \multicolumn{1}{c}{78.40}  \\
& \multicolumn{1}{c|}{EF} &  \multicolumn{1}{c}{22.90} & \multicolumn{1}{c}{96.10} & \multicolumn{1}{c}{99.20} \\
& \multicolumn{1}{c|}{WOODS} & \multicolumn{1}{c}{26.28} & \multicolumn{1}{c}{92.36} & \multicolumn{1}{c}{97.98}  \\
& \multicolumn{1}{c|}{BEOE} & \multicolumn{1}{c}{19.66} & \multicolumn{1}{c}{95.65} & \multicolumn{1}{c}{99.11}  \\
& \multicolumn{1}{c|}{TV-OOD} & \multicolumn{1}{c}{\textbf{10.53}} & \multicolumn{1}{c}{\textbf{97.18}} & \multicolumn{1}{c}{\textbf{99.26}} \\
\hline
\end{tabularx}

\par }
\end{minipage}\hfill
\begin{minipage}{.46\linewidth}
\vspace{1cm}
\subcaption{Results on ViT-B\_16 using ImageNet-1k as $D_{in}$.}
\fontsize{6}{7.2}\selectfont{\begin{tabularx}{\columnwidth}{XX|XXX}
\hline
\multicolumn{2}{c|}{\centering  $D_{in}$~+~model} &\multicolumn{3}{c}{\centering  ImageNet-1k~+~ViT-B\_16}    \\
\hline 
\multicolumn{2}{c|}{} & \multicolumn{1}{c}{ FPR95$\downarrow$} & \multicolumn{1}{c}{AUROC$\uparrow$} & \multicolumn{1}{c}{ AUPR$\uparrow$} \\
\hline
\multicolumn{1}{c}{\multirow{9}{*}{\makecell{w/o \\ auxiliary \\ OOD \\ dataset}}} & \multicolumn{1}{c|}{MSP} & \multicolumn{1}{c}{48.09} & \multicolumn{1}{c}{87.41} & \multicolumn{1}{c}{97.29}  \\
& \multicolumn{1}{c|}{Energy} & \multicolumn{1}{c}{32.31} & \multicolumn{1}{c}{93.00} & \multicolumn{1}{c}{98.52}  \\
& \multicolumn{1}{c|}{ODIN} & \multicolumn{1}{c}{34.45} & \multicolumn{1}{c}{92.58} & \multicolumn{1}{c}{98.44}  \\
& \multicolumn{1}{c|}{Mahalanobis} & \multicolumn{1}{c}{37.50} & \multicolumn{1}{c}{91.28} & \multicolumn{1}{c}{98.12} \\
& \multicolumn{1}{c|}{LogitNorm} & \multicolumn{1}{c}{\textbf{28.41}} & \multicolumn{1}{c}{\textbf{93.39}} & \multicolumn{1}{c}{98.69} \\
& \multicolumn{1}{c|}{KNN} & \multicolumn{1}{c}{29.31} & \multicolumn{1}{c}{92.58} & \multicolumn{1}{c}{98.59} \\
& \multicolumn{1}{c|}{ASH-S} & \multicolumn{1}{c}{34.17} & \multicolumn{1}{c}{82.35} & \multicolumn{1}{c}{98.35} \\
& \multicolumn{1}{c|}{VIM} & \multicolumn{1}{c}{31.13} & \multicolumn{1}{c}{93.07} & \multicolumn{1}{c}{98.57} \\
& \multicolumn{1}{c|}{TV-OOD} & \multicolumn{1}{c}{30.01} & \multicolumn{1}{c}{92.34} & \multicolumn{1}{c}{\textbf{98.75}} \\
\hline
\multicolumn{1}{c}{\multirow{5}{*}{\makecell{with \\ auxiliary \\ OOD \\ dataset}}} & \multicolumn{1}{c|}{OE} & \multicolumn{1}{c}{22.97} & \multicolumn{1}{c}{\textbf{95.00}} & \multicolumn{1}{c}{81.35}  \\
& \multicolumn{1}{c|}{EF} & \multicolumn{1}{c}{26.58} & \multicolumn{1}{c}{92.69} & \multicolumn{1}{c}{98.22} \\
& \multicolumn{1}{c|}{WOODS} & \multicolumn{1}{c}{22.95} & \multicolumn{1}{c}{93.38} & \multicolumn{1}{c}{98.56}  \\
& \multicolumn{1}{c|}{BEOE} & \multicolumn{1}{c}{25.31} & \multicolumn{1}{c}{93.86} & \multicolumn{1}{c}{98.66}  \\
& \multicolumn{1}{c|}{TV-OOD} & \multicolumn{1}{c}{\textbf{22.28}} & \multicolumn{1}{c}{94.81} & \multicolumn{1}{c}{\textbf{98.77}}  \\
\hline

\end{tabularx}}
\end{minipage}
    
\end{table*}


\subsection{Experimental Configuration}

The baseline methods are introduced in \cref{sec:related}. We use CIFAR-100 and ImageNet-1k as ID data ($D_{in}$), divided into training ($D_{in}^{train}$) and test sets ($D_{in}^{test}$). For CIFAR-100 training, we utilize 80 Million Tiny Images, excluding overlaps with CIFAR datasets. For ImageNet-1k training, we use OpenImage-O and ImageNet-O, avoiding overlap. During testing, CIFAR-100 utilize Textures, SVHN, Places365, LSUN variants, and iSUN as OOD test datasets. ImageNet-1k employs DTD, iNaturalist, Places365, and SUN for testing.

We use DenseNet121, WideResNet, and ViT-B\_16 for CIFAR datasets, and ViT-B\_16 for ImageNet-1k. For baselines, we use hyperparameters from the authors’ implementations when available, or test at least 5 hyperparameter sets otherwise. The hyperparameter $m$ is set to 0.5 across all experiments. This value was fixed based on initial validation to demonstrate consistent performance across all architectures and datasets.

\subsection{Principal Findings}

Results in \cref{table:1} show that TV-OOD consistently matches or outperforms baselines across various datasets and architectures, both with and without auxiliary OOD data.

\subsection{Effectiveness of the Generated Auxiliary OOD Dataset $D_{aug}$}

We evaluate the auxiliary OOD dataset $D_{aug}$ by training BEOE and TV-OOD with and without it. Additionally, we test $D_{AugMix}$ and $D_{CutMix}$, created by augmenting ID data with AugMix\cite{augmix} or CutMix\cite{cutmix}, followed by $D_{aug}$’s generation process, and $D_{OpenGAN}$, containing synthetic negatives from OpenGAN\cite{opengan}. \cref{table:2} shows that $D_{aug}$ enhances both methods on DenseNet121, while $D_{AugMix}$, $D_{CutMix}$, and $D_{OpenGAN}$ occasionally yield the best results, indicating room for improving OOD dataset generation.

\begin{table}[htbp]

    \centering
    \caption{Effectiveness of $D_{aug}$}
    \label{table:2}
    \scriptsize 

        \fontsize{7}{8.4}\selectfont{\begin{tabularx}{\columnwidth}{X|XXXX}
\hline
\multicolumn{1}{c|}{\centering $D_{in}$} & \multicolumn{4}{c}{cifar100} \\
\hline
\multicolumn{1}{c|}{\centering method} & \multicolumn{1}{c|}{\centering OOD datasets} & \multicolumn{1}{c}{FPR95$\downarrow$} & \multicolumn{1}{c}{AUROC$\uparrow$} & \multicolumn{1}{c}{AUPR$\uparrow$} \\
\hline
\multirow{4}{*}{BEOE} & \multicolumn{1}{l|}{$D_{out}^{train}$} & 47.58 & 91.35 & 98.01 \\

 & \multicolumn{1}{l|}{$D_{out}^{train},D_{aug}$} & 32.50 & \textbf{93.19} & 98.50 \\
 & \multicolumn{1}{l|}{$D_{out}^{train},D_{AugMix}$} & 31.50 & 92.89 & 98.35 \\
 & \multicolumn{1}{l|}{$D_{out}^{train},D_{CutMix}$}& 37.47 & 91.32 & 97.92 \\
 & \multicolumn{1}{l|}{$D_{out}^{train},D_{OpenGAN}$}& \textbf{30.82} & 92.90 & \textbf{98.61} \\
\hline
\multirow{4}{*}{TV-OOD} & \multicolumn{1}{l|}{$D_{out}^{train}$} & 44.45 & 90.95 & 97.89 \\

 & \multicolumn{1}{l|}{$D_{out}^{train},D_{aug}$} & \textbf{24.93} & 93.87 & 98.42 \\

 & \multicolumn{1}{l|}{$D_{out}^{train},D_{AugMix}$} & 27.21 & 93.59 & 98.41 \\
 
 & \multicolumn{1}{l|}{$D_{out}^{train},D_{CutMix}$} & 33.02 & 92.27 & 98.10 \\
 & \multicolumn{1}{l|}{$D_{out}^{train},D_{OpenGAN}$} & 25.75 & \textbf{94.77} & \textbf{98.76} \\

\hline
\end{tabularx} \par}\
\end{table}

\subsection{Comparison with Other f-divergences}
\label{sec:f-d}

We compare total variation (TV) with various f-divergences to highlight its advantages, including the Donsker-Varadhan (DV) representation and Nguyen's KL divergence, Reverse KL (RKL), Pearson’s $\chi^2$ (P-$\chi^2$), Squared Hellinger (SH), Jensen-Shannon (JS), and the GAN discriminator metric\cite{GAN}. DV incorporates an exponential moving average to reduce bias. Additionally, we evaluate against a binary classifier (BC) using a binary cross-entropy objective for IN/OOD classification. Unlike BC, whose logits lack probabilistic interpretation\cite{logitnorm}, our approach interprets the neural network as an f-divergence conjugate, rooted in information theory. Results are shown in \cref{f-divergences}. While TV-OOD shows strong empirical results, future work will further explore the theoretical foundations making total variation uniquely suited for OOD detection.

\begin{table}[tbp]
    \centering
    
        \centering
        \caption{FPR95 values for WideResNet using CIFAR-100 as $D_{in}$ and ViT-B\_16 using ImageNet-1k as $D_{in}$, evaluated with various f-divergences and a binary classifier. Smaller values indicate better performance.}
         \label{f-divergences}
        \fontsize{6.2}{6.2}\selectfont{
        \newcolumntype{Y}{>{\centering\arraybackslash}X}
        \setlength{\tabcolsep}{0.5mm}
        \begin{tabularx}{\linewidth}{YYYYYYYYY}

\hline
& \multicolumn{8}{c}{\centering CIFAR-100}    \\
\hline
   TV & DV & Nguyen & RKL & P-$\chi^2$ & SH & JS & GAN &  BC \\

  \textbf{41.58} & 52.92 & 55.24 & 51.75 & 92.31 & 53.03 & 52.15 & 52.44 & 57.59 \\
\hline
& \multicolumn{8}{c}{\centering ImageNet-1k}    \\
\hline
 TV & DV & Nguyen & RKL & P-$\chi^2$ & SH & JS & GAN &  BC \\
 \textbf{30.01} & 41.78 & 44.73 & 37.12 & 82.11 & 38.09 & 40.52 & 36.25 & 45.62 \\
\hline
\end{tabularx}}
   
\end{table}

\section{Conclusion}

We introduced TV-OOD, the first method using total variation for OOD detection in image classification. By modifying the MINE process to use total variation instead of KL divergence, our approach shows strong theoretical and empirical results. This work suggests that integrating information theory could offer new insights in the field of OOD detection.

\vfill\pagebreak

\bibliographystyle{IEEEbib}
\bibliography{strings,refs}

\end{document}